\documentclass[letterpaper]{article}
\usepackage{aaai23-r2hcai}  
\usepackage{times}
\usepackage{helvet}
\usepackage{courier}
\usepackage[hyphens]{url}
\usepackage{graphicx}
\usepackage{natbib}
\usepackage{caption}
\usepackage{hyperref}    
\usepackage{algorithm}
\usepackage{algorithmic}
\usepackage{amsmath, amssymb}
\usepackage{booktabs}
\usepackage{multirow}
\usepackage{float}
\usepackage{stfloats}
\usepackage[table]{xcolor}
\usepackage{subfig}  
\usepackage{tabularx}
\usepackage{xcolor}
\usepackage[T1]{fontenc}
\usepackage[utf8]{inputenc} 
\usepackage{lmodern} 
\usepackage{graphicx}
\usepackage{array}

\title{{\LARGE \bfseries Cross-LoRA: A Data-Free LoRA Transfer Framework across Heterogeneous LLMs}}

\author {
    \normalsize
    Feifan Xia\textsuperscript{1,2}, 
    Mingyang Liao\textsuperscript{1,3}, 
    Yuyang Fang\textsuperscript{4}, 
    Defang Li\textsuperscript{1}, 
    Yantong Xie\textsuperscript{1,5}, 
    Weikang Li\textsuperscript{3}, 
    Yang Li\textsuperscript{1}\thanks{Corresponding author}, 
    Deguo Xia\textsuperscript{1}, 
    Jizhou Huang\textsuperscript{1}
}

\affiliations {
    \textsuperscript{1}Baidu Inc \\
    \textsuperscript{2}Imperial College London \quad
    \textsuperscript{3}Peking University \\
    \textsuperscript{4}Zhejiang University \quad
    \textsuperscript{5}Carnegie Mellon University \\
    \texttt{\{xiafeifan, liaomingyang, lidefang, xieyantong, liyang164, xiadeguo, huangjizhou01\}@baidu.com} \\
    \texttt{feifan.xia23@imperial.ac.uk, 2301210317@stu.pku.edu.cn, wavejkd@pku.edu.cn,} \\
    \texttt{fangyuyang@zju.edu.cn, yantongx@andrew.cmu.edu}
}

\begin{document}
\nocopyright   
\maketitle

\begin{abstract}
Traditional parameter-efficient fine-tuning (PEFT) methods such as \textbf{LoRA} are tightly coupled with the base model architecture, which constrains their applicability across heterogeneous pretrained large language models (LLMs). To address this limitation, we introduce \textbf{Cross-LoRA}, a \textbf{data-free} framework for transferring LoRA modules between diverse base models without requiring additional training data. \textbf{Cross-LoRA} consists of two key components: \textbf{(a) LoRA-Align}, which performs subspace alignment between source and target base models through rank-truncated singular value decomposition (SVD) and Frobenius-optimal linear transformation, ensuring compatibility under dimension mismatch; and \textbf{(b) LoRA-Shift}, which applies the aligned subspaces to project source LoRA weight updates into the target model parameter space. Both components are \textbf{data-free}, \textbf{training-free}, and enable lightweight adaptation on a commodity GPU in 20 minutes. Experiments on ARCs, OBQA and HellaSwag show that \textbf{Cross-LoRA achieves relative gains of up to 5.26\% over base models}. Across other commonsense reasoning benchmarks, Cross-LoRA maintains performance comparable to that of directly trained LoRA adapters.
\end{abstract}

\section{Introduction}

Large Foundation Models (LFMs) such as GPT‑4, LLaMA‑3, Qwen2.5, and Gemma‑2 have become the cornerstone of modern artificial intelligence, delivering state‑of‑the‑art performance across diverse domains, particularly natural language processing ~\cite{openai_gpt-4_2024,qwen_qwen25_2025,team_gemma_2024}. Owing to their strong capabilities, fine-tuning LFMs for downstream tasks has become the prevalent strategy. However, conventional fine-tuning generally requires saving a new checkpoint of comparable size to the original model for each task, resulting in substantial storage demands and memory overhead as model scale and task diversity increase.

\begin{figure}[H]
    \centering
    \includegraphics[width=1\linewidth]{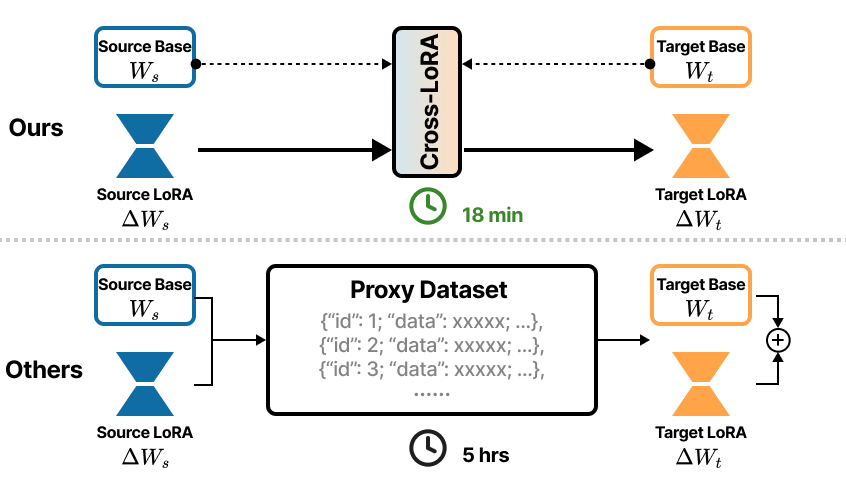}
    \caption{Overview of the Cross-LoRA framework.
    The figure illustrates the process of transferring a LoRA adapter from a source model to a target model.
    It highlights the rank-truncated SVD decomposition of source and target bases, subspace alignment via Frobenius-optimal projection, 
    and the generation of target-compatible LoRA updates.}
    \label{fig:method_overview}
\end{figure}

To address this challenge, Parameter-Efficient Fine-Tuning (PEFT) ~\cite{xu_parameter-efficient_2023} methods have been widely adopted. Among them, Low-Rank Adaptation (LoRA) ~\cite{hu_lora_2021}, DoRA~\cite{liu_dora_2024}, and LoRA+ ~\cite{hayou_lora_2024} are particularly effective, representing weight updates as low-rank matrices to reduce the number of trainable parameters. These designs achieve strong task performance with relatively modest computational cost.

Nevertheless, LoRA suffers from a key limitation: its dependency on the base model. A LoRA adapter fine-tuned on one model cannot be directly applied to another without retraining. This issue becomes especially problematic when the base model is replaced or deprecated, as the associated LoRA adapters must either be retrained or discarded. Retraining is often infeasible due to the unavailability of the original fine-tuning data and the high computational cost of additional tuning.

This raises a crucial question:
\vspace{0.5em}
\begin{center}
  \begin{tabular}{|p{0.4\textwidth}|}
    \hline
    \textbf{Question:} Can we transfer LoRA adapters trained on one base model to another, without requiring any training data or additional fine-tuning? \\
    \hline
  \end{tabular}
\end{center}
\vspace{0.5em}

We propose \textbf{Cross-LoRA}, a data-free and training-free framework for transferring LoRA adapters across heterogeneous large language models. Cross-LoRA introduces two key components: LoRA-Align, which performs rank-truncated SVD–based subspace alignment and Frobenius-optimal linear transformation to ensure compatibility across dimension mismatches; and LoRA-Shift, which projects the aligned source LoRA weight updates into the target model parameter space.\\

Our contributions can be summarized as follows:\\
\begin{enumerate}
    \item Data-free and training-free LoRA transfer at LLM scale. To our knowledge, Cross-LoRA is the first framework enabling LoRA migration across heterogeneous large language models without requiring access to the original training data or additional fine-tuning.

    \item Efficient alignment mechanism tailored for LLMs. We introduce a Frobenius-optimal subspace alignment approach with rank-truncated singular value decomposition (SVD), which ensures numerical stability under dimension mismatch and enables practical, lightweight transfer on autoregressive LLMs without the need for costly computation or retraining.
    \item Robust generalization across architectures and model families. Cross-LoRA demonstrates strong transferability across heterogeneous base models, such as Qwen, LLaMA, and Gemma. It is architecture agnostic and plug and play, requiring no modification to the tokenizer or the internal structure of the base model.

\end{enumerate}

\section{Related Work}

Transferring knowledge between models has long been a central goal in machine learning. 
A common approach is Knowledge Distillation \cite{hinton_distilling_2015}, 
where a smaller “student” learns to mimic a more powerful “teacher.” 
Later studies examined its convergence and generalization \cite{phuong_towards_2021, kaplun_knowledge_2022}. 
The student acquires the teacher’s behavior through training on this dataset, 
thus obtaining \( W_0 \) or a \( W_0 \)-equivalent when base models differ.

Using original or synthetic data as a proxy for transferring delta weights has also shown viability. Trans-LoRA~\cite{wang_trans-lora_2025} generates synthetic data from the source model combined with a trained LoRA, then uses this data to perform LoRA training on the target model. This method has proven effective across model families, yet proxy data may fail to represent the original distribution, and the transfer process requires training time comparable to fine-tuning the original LoRA.

To reduce reliance on data and training, LoRASuite~\cite{li_lorasuite_2025} compares subspaces between the source and target models and constructs a transfer matrix, followed by lightweight fine-tuning on the target. While efficient, it is limited to same-architecture transfers (e.g., small-to-large within a model family) and still requires partial access to target model gradients.

In contrast, LoRA-X~\cite{farhadzadeh_lora-x_2025} avoids both data dependency and training. It proposes a subspace-constrained alternative to LoRA that can align with the subspaces across different models. However, it requires specially designed adapters and has thus far been primarily applied to diffusion models. 

ProLoRA~\cite{farhadzadeh_zero-shot_2025} further removes this constraint by enabling training-free transfer of standard LoRA adapters. It decomposes LoRA weights into subspace and null-space components, projecting them into the target model’s structure without any data or training. Though more general, its application remains focused on text-to-image diffusion models.

In summary, existing methods either rely on proxy data and training(Trans-LoRA, LoRASuite), or are adapter- or domain-specific (LoRA-X, ProLoRA). We aim to bridge this gap with a general, training-free transfer method for standard LoRA adapters, applicable to Large Language Models.

\section{Method}
\begin{figure*}[t]
\centering
\includegraphics[width=0.8\textwidth]{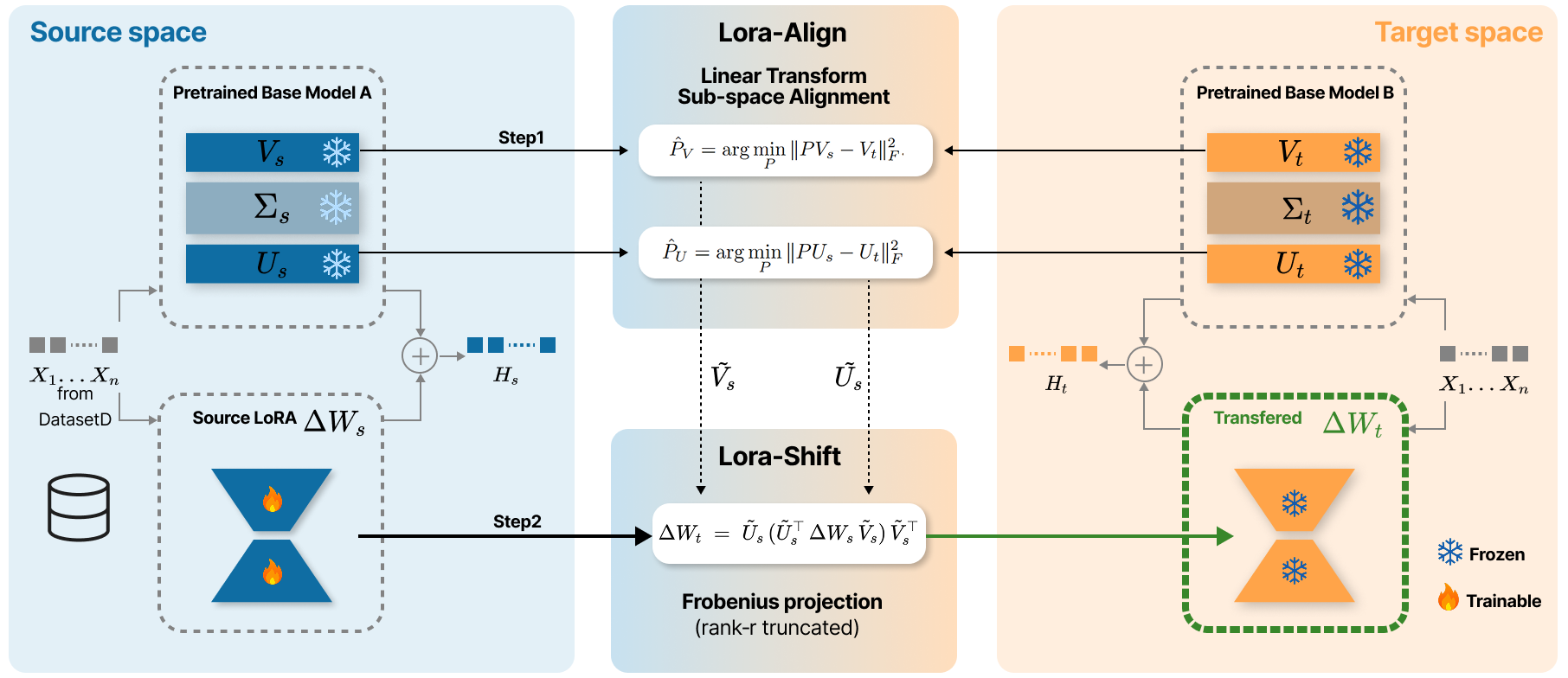} 
\caption{Overall framework of Cross-LoRA. The source base model (left) provides
        LoRA updates $\Delta W_s$ trained on a specific task. 
        In Step 1, rank-truncated SVD is applied to both source and target base weights 
        to extract compact subspaces. Lora-Align computes Frobenius-optimal linear transformations 
        to align the source subspaces with the target. 
        In Step 2, Lora-Shift projects the source LoRA updates into the aligned target subspace, 
        yielding target-compatible LoRA weights $\Delta W_t$.
        The process requires no access to training data and runs efficiently on commodity GPUs.}
\label{fig2}
\end{figure*}
\subsection{Problem Formulation}
Traditional parameter-efficient fine-tuning (PEFT) methods such as LoRA are highly effective but intrinsically tied to the base model, preventing direct reuse across heterogeneous architectures. Inspired by LoRA-X~\cite{farhadzadeh_lora-x_2025}, which demonstrated data-free transfer for diffusion models with structurally similar backbones, we extend this idea to heterogeneous large language models (LLMs) and propose a general framework for transferring LoRA modules without access to training data or further optimization.

Formally, let the source base model weight be 
$W_s \in \mathbb{R}^{m \times n}$ 
and the corresponding LoRA update be 
$\Delta W_s \in \mathbb{R}^{m \times n}$.
Given a target base model weight 
$W_t \in \mathbb{R}^{m' \times n'}$, 
our objective is to construct a target-compatible LoRA update 
$\Delta W_t \in \mathbb{R}^{m' \times n'}$ 
such that the adapted target model retains the knowledge encoded by $\Delta W_s$, without requiring original data or further training.

\vspace{0.5em}

\subsection{Framework Overview: Cross-LoRA}
Cross-LoRA consists of two complementary components:

\textbf{LoRA-Align.}  
We perform subspace alignment between source and target base models using rank-truncated singular value decomposition (SVD). This resolves dimensional mismatches and identifies a shared subspace for LoRA transfer.

\textbf{LoRA-Shift.}  
We project the source LoRA update $\Delta W_s$ into the aligned subspace to obtain a target-compatible update $\Delta W_t$. This transformation is both data-free and training-free.

\vspace{0.5em}

\subsection{Mathematical Formulation}
\paragraph{SVD Decomposition.}
We first compute rank-$r$ truncated SVDs for the source and target base weights:
\begin{equation}
W_s \approx U_s \Sigma_s V_s^\top
\end{equation}
\begin{equation}
W_t \approx U_t \Sigma_t V_t^\top
\end{equation}
where $U_s \in \mathbb{R}^{m \times r}$, $V_s \in \mathbb{R}^{n \times r}$, and similarly for $U_t, V_t$. The truncated singular vectors capture the dominant directions of the parameter spaces.

\vspace{0.5em}

\paragraph{Subspace Alignment.}
To establish subspace compatibility, we construct Frobenius-optimal linear transforms that align the source and target bases:
\begin{equation}
\hat{P}_U = \arg\min_P \|PU_s - U_t\|_F^2
\end{equation}
\begin{equation}
\hat{P}_V = \arg\min_P \|PV_s - V_t\|_F^2
\end{equation}
These least-squares problems admit closed-form solutions and remain stable even when $U_s$ or $V_s$ is rank-deficient. In practice, we solve them using efficient batched routines (e.g., \texttt{torch.linalg.lstsq}). The aligned subspaces are then defined as:
\begin{equation}
\tilde{U}_s = \hat{P}_U U_s
\end{equation}
\begin{equation}
\tilde{V}_s = \hat{P}_V V_s
\end{equation}

\vspace{0.5em}

\paragraph{Frobenius Projection.}
We then project the source LoRA update $\Delta W_s$ into the aligned target subspace:
\begin{equation}
\Delta W_t = \tilde{U}_s \left(\tilde{U}_s^\top \Delta W_s \tilde{V}_s \right) \tilde{V}_s^\top
\end{equation}
This operation minimizes the Frobenius norm $\|\Delta W_s - \Delta W_t\|_F$ under the aligned latent basis, enabling adaptation to the target model geometry without requiring training data or explicit pseudo-inverses.

\vspace{0.5em}

\subsection{Complexity and Efficiency}

\begin{algorithm}[ht]
\caption{Cross-LoRA Transfer via Subspace Projection}
\label{alg:crosslora}
\textbf{Input}: LoRA update $\Delta W_s$, source weights $W_s$, target weights $W_t$ \\
\textbf{Parameter}: Truncated rank $r$ \\
\textbf{Output}: Transferred update $\Delta W_t$
\begin{algorithmic}[1]
\STATE Initialize empty $\Delta W_t$ and counters.
\FOR{each LoRA parameter $k$ in $\Delta W_s$}
    \STATE Determine base key $b$ from $k$.
    \IF{$b \notin W_s$ or $b \notin W_t$}
        \STATE \textbf{continue}
    \ENDIF
    \STATE Compute rank-$r$ SVDs for $W_s[b]$, $W_t[b]$.
    \STATE Derive aligned basis $\tilde{U}_s, \tilde{V}_s$ via least-squares.
    \IF{$k$ is a left LoRA weight}
        \STATE Project: $\Delta W_t[k] \gets \tilde{U}_s (\tilde{U}_s^\top \Delta W_s[k])$
    \ELSE
        \STATE Project: $\Delta W_t[k] \gets (\Delta W_s[k] \tilde{V}_s) \tilde{V}_s^\top$
    \ENDIF
    \STATE Cast to FP16 and update statistics.
\ENDFOR
\STATE \textbf{return} $\Delta W_t$
\end{algorithmic}
\end{algorithm}

We analyze the computational complexity and accuracy implications 
of rank-truncated SVD in Cross-LoRA. 

\vspace{0.5em}

Let the base weight 
\[
W_0 \in \mathbb{R}^{m \times n}
\] 
have singular value decomposition (SVD):
\[
W_0 = U \Sigma V^\top,
\]
with truncated form
\[
W_0^{(r)} = U_r \Sigma_r V_r^\top.
\]

\vspace{0.8em}

By the Eckart--Young--Mirsky theorem~\cite{mirsky_symmetric_1960}, 
$W_0^{(r)}$ is the best rank-$r$ approximation of $W_0$ under the Frobenius norm.

\vspace{0.8em}

Since a LoRA update can be written as
\[
\Delta W_s = B_s A_s,
\]
with $\mathrm{rank}(\Delta W_s) \leq r$, 
it can be expressed in the truncated subspace as
\[
\Delta W_s \approx U_r C V_r^\top, 
\quad C \in \mathbb{R}^{r \times r}.
\]

\vspace{0.8em}

Projecting into the truncated subspace gives
\[
\widehat{\Delta W_s} 
   = U_r \big(U_r^\top \Delta W_s V_r\big) V_r^\top.
\]

The resulting error is
\[
E = \|\Delta W_s - \widehat{\Delta W_s}\|_F 
   \;\; \approx 0,
\]
since discarded singular directions contribute negligibly 
when $r$ is chosen at least as large as the LoRA rank 
(typically $16$--$64$).

\vspace{1em} 

Truncation reduces the time complexity from 
\[
O(mn \cdot \min(m,n)) 
\quad \text{(full SVD)}
\]
to
\[
O(rmn)
\quad \text{(truncated SVD)},
\]
and memory cost from 
\[
O(mn) \quad \text{to} \quad O(r(m+n)).
\]

\vspace{0.8em}

This efficiency enables transfer on a single $8$GB GPU 
within $20$ minutes.  
Previous spectral studies show that the first $r=320$ singular values capture over $99\%$ of the Frobenius norm energy.

\vspace{0.8em}
\begin{table}[H]
\centering
\small
\setlength{\tabcolsep}{4pt}
\resizebox{\columnwidth}{!}{%
\begin{tabular}{lccc}
\toprule
Method & Time & Memory & F.N. Energy \\
\midrule
Full-rank & $O(mn\!\cdot\!\min)$ & $O(mn)$ & 100\% \\
Truncated ($r\!\ll\!\min$) & $O(rmn)$ & $O(r(m+n))$ & $\geq$99\% \\
\bottomrule
\end{tabular}
}
\caption{Computational and representational trade-off between full-rank and rank-truncated SVD in Cross-LoRA. 
Truncation substantially reduces time and memory complexity while retaining at least 99\% of the Frobenius norm energy, 
as detailed in \textbf{Appendix~C}.}
\label{tab:truncation}
\end{table}

\section{Experiments}
\vspace{0.8em}
\begin{table*}[tb]
\centering
\small
\setlength{\tabcolsep}{5pt}
\resizebox{\textwidth}{!}{%
\begin{tabular}{l l c c c c c}
\toprule
\textbf{Model} & \textbf{Adapter} & \textbf{Training} & \textbf{Arc-c (Acc.)} & \textbf{Arc-e (Acc.)} & \textbf{OpenBookQA (Acc.)} & \textbf{HellaSwag (Acc.)} \\
\midrule
\multirow{3}{*}{LLaMA-3.2-3B} 
& Base Model        & No  & 0.6838 & 0.8260 & 0.7260 & 0.5657 \\
& Trained LoRA      & Yes & \textbf{0.7082} (+3.57\%)  & 0.8229 (-0.38\%) & 0.7200 (-0.83\%) & \textbf{0.6092} (+7.69\%) \\
\rowcolor{gray!15} & Transferred LoRA  & No  &\textbf{0.7065 (+3.32\%)}& \textbf{0.8331 (+0.86\%)} & 0.7260 (+0.00\%) & 0.5667 (+0.18\%) \\
\midrule
\multirow{3}{*}{Gemma-2-2B}
& Base Model        & No  & 0.7144 & 0.8269 & 0.7120 & 0.5254 \\
& Trained LoRA      & Yes & 0.7355 (+2.95\%) & 0.8345 (+0.92\%) & \textbf{0.7220} (+1.40\%) & 0.5100 (-2.93\%) \\
\rowcolor{gray!15} & Transferred LoRA  & No  & \textbf{0.7520 (+5.26\%)} & 0.8289 (+0.24\%) & 0.7120 (+0.00\%) & \textbf{0.5254} (+0.00\%)  \\
\midrule
\multirow{3}{*}{Qwen2.5-3B}
& Base Model        & No  & 0.8129 & 0.9061 & 0.8000 & 0.6040 \\
& Trained LoRA      & Yes & \textbf{0.8172} (+0.53\%)$^{\dagger}$ & 0.9061 (+0.00\%) & \textbf{0.8020} (+0.25\%)$^{\dagger}$ & \textbf{0.6125} (+1.41\%)$^{\dagger}$ \\
\rowcolor{gray!15} & Transferred LoRA  & No  & 0.8130 (+0.01\%) & \textbf{0.9063}  (+0.02\%)$^{\dagger}$ & 0.8000 (+0.00\%) & 0.6028 (-0.20\%) \\
\midrule
\multirow{3}{*}{Qwen2.5-1.5B}
& Base Model        & No  & 0.7338 & 0.8483 & 0.7440 & 0.4980 \\
& Trained LoRA      & Yes &\textbf{0.7457} (+1.62\%)  & 0.8485 (+0.02\%) & \textbf{0.7500} (+0.81\%) & \textbf{0.5180} (+4.02\%) \\
\rowcolor{gray!15} & Transferred LoRA  & No  & 0.7372 (+0.46\%) & \textbf{0.8485} (+0.02\%) & 0.7460 (+0.27\%) & 0.4995 (+0.30\%) \\
\bottomrule
\end{tabular}
}
\caption{Performance comparison of Cross-LoRA with base models and directly trained LoRA adapters across four benchmarks. 
Transferred adapters are data-free and training-free. The best results per dataset are marked with $^{\dagger}$. Percentage change is relative to the corresponding base model.}
\label{tab:crosslora-results}
\end{table*}

This section evaluates the effectiveness of Cross-LoRA for training-free LoRA transfer across heterogeneous large language models (LLMs). We first describe the experimental setup and then present results on multiple NLP benchmarks.

\subsection{Experiment Setup}
We evaluate the effectiveness of Cross-LoRA across multiple large language models (LLMs) and benchmark datasets. 
Our goal is to assess whether Cross-LoRA can transfer LoRA adapters trained on a source model to a heterogeneous target model 
without requiring any additional training or access to the original fine-tuning data.

\paragraph{Models.}
We select four representative instruction-tuned LLMs as the base models for our experiments:
LLaMA-3.2-3B, Qwen2.5-1.5B, Qwen2.5-3B, and Gemma-2-2B. 
These models cover both decoder-only architectures and different model families, 
enabling evaluation of transferability across heterogeneous base models.

\paragraph{Datasets.}
We benchmark performance on four widely used NLP tasks: 
ARC-c, ARC-e, OpenBookQA, and HellaSwag~\cite{clark_think_2018,mihaylov_can_2018,zellers_hellaswag_2019}. 
These tasks are standard in evaluating reasoning, knowledge retrieval, and commonsense understanding, 
providing a comprehensive view of model performance under transfer.

\paragraph{Baselines.}
We compare three settings:
(1) the \emph{Base Model}, evaluated without any LoRA adapters;
(2) \emph{Trained LoRA}, where adapters are directly fine-tuned on the target model for each dataset. This training process follows common practices in prior PEFT transfer studies~\cite{liang_drag-and-drop_2025};
and (3) \emph{Transferred LoRA} (Cross-LoRA), where adapters fine-tuned on a source model are transferred to the target model using our framework, 
without any training data or fine-tuning. 
This comparison allows us to quantify the relative utility of Cross-LoRA, highlighting both its advantages and limitations in zero-shot adaptation settings.

\paragraph{Implementation.}
Cross-LoRA employs rank-truncated SVD and Frobenius-optimal subspace alignment to map LoRA weight updates into the target model space.
All experiments are executed on a single NVIDIA V100 GPU.  
No additional training or optimization is performed on the transferred adapters. A detailed description of the setup is provided in Appendix~A.

\subsection{Main Results}

Table~\ref{tab:crosslora-results} presents the results of Cross-LoRA across four QA benchmarks. We compare transferred adapters against base models and directly trained LoRA adapters, reporting accuracy and relative gains. Overall, transferred adapters yield consistent improvements over base models, achieving an average relative gain of \textbf{+0.848\%}, closely approaching the \textbf{+0.976\%} gain from trained LoRA. Notably, the best observed gains are \textbf{+5.26\% on ARC-c} and \textbf{+3.32\% on ARC-e}, demonstrating the effectiveness of training-free subspace alignment in recovering task-specific knowledge.

Compared to trained LoRA, transferred adapters deliver competitive results. On ARC-e and OpenBookQA, Cross-LoRA matches or slightly exceeds the trained counterpart on multiple architectures. On HellaSwag, performance varies more across models, with some configurations showing minor drops. This variation suggests architecture-dependent transfer fidelity, especially in tasks requiring fine-grained reasoning.

These findings indicate that Cross-LoRA enables efficient adapter transfer in both homogeneous and heterogeneous settings, offering a practical alternative when fine-tuning data or computational resources are limited. While minor performance gaps remain in specific tasks, the results suggest room for further refinement, such as lightweight post-transfer adaptation.

\subsection{Effect of Subspace Alignment}
To further understand the contribution of subspace alignment, we compare Cross-LoRA against a linear interpolation baseline, as well as base and directly trained LoRA adapters. 
The results, reported in Table~\ref{tab:alignment}, are obtained when transferring a LoRA trained on Qwen2.5-1.5B into LLaMA-3.2-3B. 
The interpolation baseline adjusts LoRA weights through direct dimension matching without subspace projection, 
while Cross-LoRA applies rank-truncated SVD and Frobenius-optimal alignment. 

The results demonstrate that simple interpolation provides little benefit and in some cases degrades performance relative to the base model. 
By contrast, Cross-LoRA consistently improves accuracy on ARC-c and ARC-e, reaching 0.7065 and 0.8331 respectively, which is close to or above the directly trained LoRA. 
The gain is particularly evident compared to interpolation, highlighting the necessity of subspace alignment in capturing transferable task-specific knowledge. 
On HellaSwag, both methods achieve only marginal improvements over the base model, suggesting that the task requires finer-grained reasoning that may not be fully preserved by projection alone. 

\begin{table}[H]
\centering
\small 
\setlength{\tabcolsep}{6pt} 
\resizebox{\columnwidth}{!}{%
\begin{tabular}{lcc}
\toprule
\textbf{Method} & \textbf{ARC-c (Acc.)} & \textbf{ARC-e (Acc.)} \\
\midrule
Base & 0.6838 & 0.8260 \\
Base + Trained LoRA & 0.7082 & 0.8229 \\
Interpolation Baseline & 0.6820 & 0.8320 \\
\rowcolor{gray!15} Cross-LoRA (Projection) & \textbf{0.7065} & \textbf{0.8331} \\
\bottomrule
\end{tabular}
}
\caption{Impact of subspace alignment when transferring from Qwen2.5-1.5B to LLaMA-3.2-3B. 
Cross-LoRA employs Frobenius projection, whereas the interpolation baseline directly resizes weights.}
\label{tab:alignment}
\end{table}

Taken together, these findings confirm that the Frobenius-optimal projection in Cross-LoRA is essential for effective transfer. 
While direct interpolation fails to preserve critical structure, subspace alignment enables the transferred LoRA to approach the performance of direct fine-tuning, 
validating the design choice of aligning singular vector subspaces during transfer.

\section{Ablation Studies}
\subsection{Rank of Cross-Lora}

\begin{figure}[ht]
    \vspace{-1pt} 
    \centering
    \subfloat[]{%
        \includegraphics[width=0.49\columnwidth]{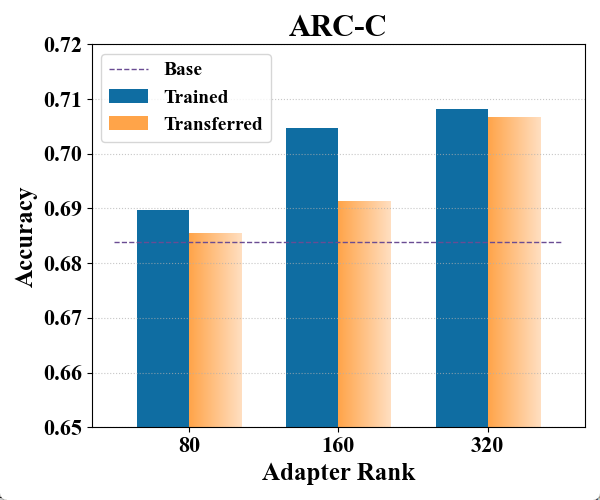}%
        \label{fig:subfig-a}}%
    \hfill
    \hspace{-5pt}
    \subfloat[]{%
        \includegraphics[width=0.49\columnwidth]{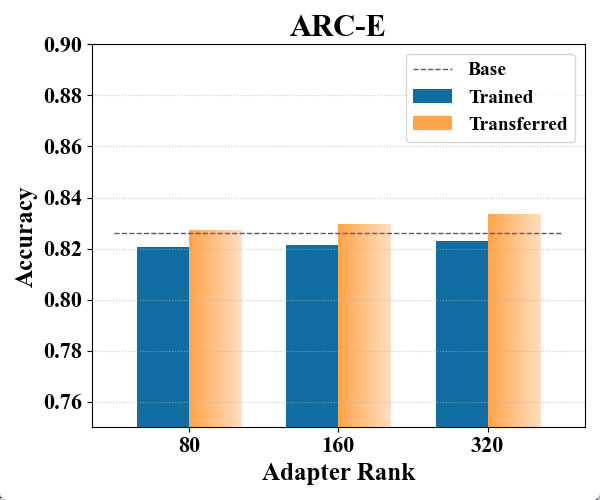}%
        \label{fig:subfig-b}}%
    \vspace{-8pt} 
    \caption{ Effect of adapter rank on Cross-LoRA performance. 
    Results on ARC-C and ARC-E demonstrate that higher adapter ranks consistently improve accuracy for both trained and transferred LoRA adapters. 
    Notably, transferred LoRA exhibits smaller performance degradation at low ranks compared to directly trained LoRA, 
    highlighting the robustness of Cross-LoRA under resource-constrained settings.}
    \vspace{-1pt} 
    \label{fig:two-subfigs}
\end{figure}

Figure~\ref{fig:two-subfigs} illustrates the effect of adapter rank on \textbf{Cross-LoRA} performance, evaluated on \textbf{ARC-C} and \textbf{ARC-E} benchmarks. We observe a consistent trend that accuracy improves as the adapter rank increases for both \textbf{trained} and \textbf{transferred LoRA}. This aligns with the intuition that \textbf{low-rank approximations restrict representational capacity}, whereas higher ranks provide richer subspaces to capture task-relevant features. However, this accuracy gain comes with increased storage cost: the \textbf{LoRA checkpoint size} grows from \textbf{3.7MB at $r=80$ to 32.4MB at $r=320$}, highlighting a trade-off between \textbf{accuracy and efficiency}.

\textbf{Notably, transferred LoRA even surpasses directly trained LoRA on the target model in some cases}, as shown in Figure~\ref{fig:subfig-b}. This phenomenon can be explained by differences in training dynamics across models. Certain target models, even under the same training recipe, may fail to converge or overfit, limiting their ability to extract effective features from the dataset. In contrast, the source model may better capture task-relevant information during LoRA fine-tuning, which Cross-LoRA can then efficiently transfer to the target. \textbf{This suggests that Cross-LoRA can leverage source models better suited to a given dataset or domain to improve weaker target models.}

At lower ranks, such as $r=80$, both trained and transferred LoRA experience performance degradation. \textbf{Notably, the decline is less severe for transferred LoRA}, indicating that the \textbf{Frobenius-optimal projection} mechanism of Cross-LoRA provides additional robustness under low-rank constraints. \textbf{This highlights a key advantage of Cross-LoRA: it maintains competitive performance even with significantly reduced adapter size}, making it particularly suitable for \textbf{resource-constrained scenarios}.

\subsection{Cross Model Transferability}

\begin{figure}[ht]
    \centering
    \includegraphics[width=1\linewidth]{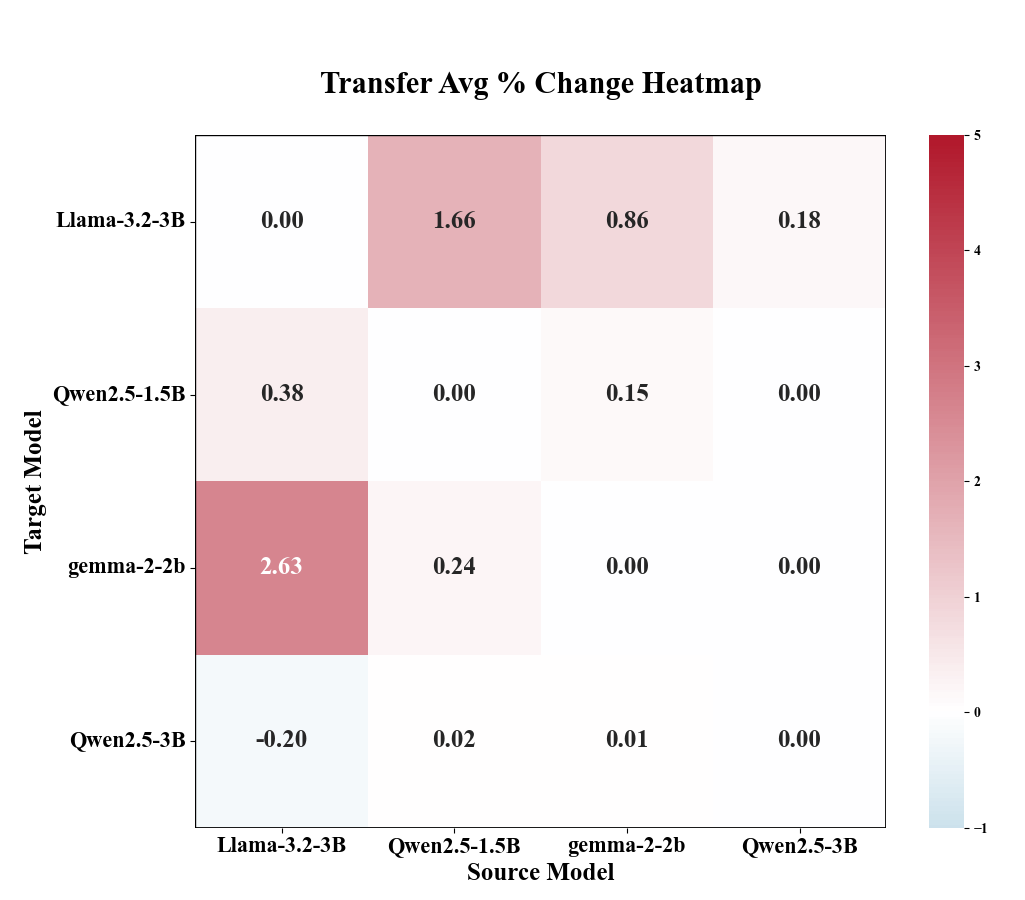}
    \caption{Cross-model transferability heatmap of Cross-LoRA. 
    Each cell reports the average percentage change in accuracy across ARC-C, ARC-E, and OpenBookQA, 
    relative to the corresponding base models. 
    Transfers between models with similar architectures (e.g., LLaMA-3.2-3B and Qwen2.5-1.5B, both using GQA and SwiGLU) yield stronger gains, 
    while transfers involving models with architectural mismatches (e.g., Gemma-2-2B with MHA and GeLU) show weaker improvements. }
    \label{fig:heatmap}
\end{figure}

To further evaluate the generalization capability of Cross-LoRA, we examine the effectiveness of transferring adapters across different large language models. Figure~\ref{fig:heatmap} reports the average percentage change relative to the corresponding base models, aggregated across ARC-C, ARC-E, and OpenBookQA.

The results reveal clear trends regarding the role of architectural similarity. Transfers between models with highly aligned configurations show consistent improvements. For instance, both LLaMA-3.2-3B and Qwen2.5 models employ grouped-query attention (GQA, 1KV), SwiGLU activation, and RMSNorm normalization. This structural alignment explains the stable positive transfer between these models, such as LLaMA-3.2-3B $\rightarrow$ Qwen2.5-1.5B (+0.38\%) and Qwen2.5-1.5B $\rightarrow$ LLaMA-3.2-3B (+1.66\%). These findings indicate that selecting \textbf{GQA + SwiGLU + RMSNorm models as source models} is particularly effective for Cross-LoRA.\\

In contrast, when transferring from Gemma-2-2B, which employs multi-head attention (MHA) with multi-KV and a mixed GeLU/SwiGLU activation scheme, the benefits are less consistent. While Gemma-2-2B $\rightarrow$ LLaMA-3.2-3B yields a notable +2.63\% improvement, reverse transfer or transfers to other targets show limited gains (e.g., Qwen2.5-3B $\rightarrow$ Gemma-2-2B at +0.01\%). This highlights that attention mechanism differences can hinder subspace alignment, limiting the robustness of the projection.\\

We also observe non-symmetric behaviors in transferability. For example, Qwen2.5-3B $\rightarrow$ LLaMA-3.2-3B exhibits a slight negative effect (-0.20\%), despite similar parameter counts. This suggests that differences in hidden layer width or attention head distributions can challenge the Frobenius projection, even when overall model scales are close.\\

Overall, these results indicate that Cross-LoRA transfer is more stable and beneficial when the source model shares the GQA, SwiGLU, and RMSNorm design with the target. This architecture-agnostic yet practically guided choice enables robust transfer performance across heterogeneous LLMs.

\section{Discussion and Future Work}

Cross-LoRA demonstrates that data-free and training-free transfer of LoRA adapters is feasible at the scale of large language models. By leveraging rank-truncated SVD and Frobenius-optimal subspace alignment, it effectively projects task-specific updates across heterogeneous architectures while maintaining efficiency on commodity GPUs. The framework consistently improves over base models and delivers performance comparable to directly trained LoRA adapters, while reducing computational and memory requirements to enable transfer on a single 8GB GPU within 20 minutes.\\

Despite these advantages, several limitations remain. First, Cross-LoRA does not fully close the gap with directly trained adapters in certain reasoning-heavy tasks, such as HellaSwag, indicating that some task-specific signals are lost during projection. Second, the effectiveness of transfer is sensitive to architectural mismatches. In particular, differences in attention mechanisms (e.g., grouped-query vs. multi-head attention) or activation functions (e.g., SwiGLU vs. ReLU) can reduce alignment quality. Our experiments indicate that source models equipped with grouped-query attention (GQA), SwiGLU activations, and RMSNorm normalization provide stronger transferability, suggesting that architectural choices substantially influence alignment effectiveness. Third, the current framework performs one-shot projection without any post-transfer adaptation, which may limit performance when the source and target distributions diverge significantly.

An additional insight from our analysis is the quantification of LoRA transferability across heterogeneous LLM architectures. Together with the plug-and-play nature of Cross-LoRA, this suggests a new paradigm: training LoRA adapters on a ``learner'' model well-suited for a given domain, and then transferring them to a ``receiver'' model for deployment, without requiring access to the original training data. This could substantially reduce the cost of LoRA-based adaptation and accelerate knowledge sharing across model families.\\

Future work could explore several directions. Hybrid approaches that combine data-free projection with lightweight task-agnostic adaptation may help recover residual task-specific performance. Extending the method to larger-scale models (e.g., 13B or 70B parameters) and multimodal architectures is another promising avenue. Automated strategies for layer-wise or subspace selection could further reduce computational overhead. Finally, a more rigorous theoretical analysis of projection error and its correlation with downstream performance would provide deeper insights into the robustness and limitations of Cross-LoRA.

\section{Conclusion}

We present Cross-LoRA, a simple yet effective framework for data-free and training-free transfer of LoRA adapters across heterogeneous language models. By aligning subspaces with truncated SVD and applying Frobenius-optimal projection, Cross-LoRA enables fast and reliable adapter migration without requiring original training data.

Experiments on multiple reasoning benchmarks show that Cross-LoRA achieves consistent improvements over base models and matches the performance of directly trained LoRA adapters, while completing transfer in minutes on a single 8GB GPU. These results highlight the practicality of Cross-LoRA for low-resource and real-world deployment scenarios.

We believe Cross-LoRA offers a scalable and lightweight alternative to conventional fine-tuning, and we hope it serves as a foundation for future research on modular, transferable, and architecture-agnostic adaptation in large language models.

\bibliographystyle{plainnat}  
\bibliography{main}

\begin{thebibliography}{23}
\providecommand{\natexlab}[1]{#1}
\providecommand{\url}[1]{\texttt{#1}}
\expandafter\ifx\csname urlstyle\endcsname\relax
  \providecommand{\doi}[1]{doi: #1}\else
  \providecommand{\doi}{doi: \begingroup \urlstyle{rm}\Url}\fi

\bibitem[Aghajanyan et~al.(2020)Aghajanyan, Zettlemoyer, and Gupta]{aghajanyan_intrinsic_2020}
Armen Aghajanyan, Luke Zettlemoyer, and Sonal Gupta.
\newblock Intrinsic {Dimensionality} {Explains} the {Effectiveness} of {Language} {Model} {Fine}-{Tuning}, December 2020.
\newblock URL \url{http://arxiv.org/abs/2012.13255}.
\newblock arXiv:2012.13255 [cs].

\bibitem[Clark et~al.(2018)Clark, Cowhey, Etzioni, Khot, Sabharwal, Schoenick, and Tafjord]{clark_think_2018}
Peter Clark, Isaac Cowhey, Oren Etzioni, Tushar Khot, Ashish Sabharwal, Carissa Schoenick, and Oyvind Tafjord.
\newblock Think you have {Solved} {Question} {Answering}? {Try} {ARC}, the {AI2} {Reasoning} {Challenge}, March 2018.
\newblock URL \url{http://arxiv.org/abs/1803.05457}.
\newblock arXiv:1803.05457 [cs].

\bibitem[Eckart and Young(1936)]{eckart_approximation_1936}
Carl Eckart and Gale Young.
\newblock The approximation of one matrix by another of lower rank.
\newblock \emph{Psychometrika}, 1\penalty0 (3):\penalty0 211--218, September 1936.
\newblock ISSN 1860-0980.
\newblock \doi{10.1007/BF02288367}.
\newblock URL \url{https://doi.org/10.1007/BF02288367}.

\bibitem[Farhadzadeh et~al.(2025{\natexlab{a}})Farhadzadeh, Das, Borse, and Porikli]{farhadzadeh_lora-x_2025}
Farzad Farhadzadeh, Debasmit Das, Shubhankar Borse, and Fatih Porikli.
\newblock {LoRA}-{X}: {Bridging} {Foundation} {Models} with {Training}-{Free} {Cross}-{Model} {Adaptation}, February 2025{\natexlab{a}}.
\newblock URL \url{http://arxiv.org/abs/2501.16559}.
\newblock arXiv:2501.16559 [cs].

\bibitem[Farhadzadeh et~al.(2025{\natexlab{b}})Farhadzadeh, Das, Borse, and Porikli]{farhadzadeh_zero-shot_2025}
Farzad Farhadzadeh, Debasmit Das, Shubhankar Borse, and Fatih Porikli.
\newblock Zero-{Shot} {Adaptation} of {Parameter}-{Efficient} {Fine}-{Tuning} in {Diffusion} {Models}, May 2025{\natexlab{b}}.
\newblock URL \url{http://arxiv.org/abs/2506.04244}.
\newblock arXiv:2506.04244 [cs].

\bibitem[Hayou et~al.(2024)Hayou, Ghosh, and Yu]{hayou_lora_2024}
Soufiane Hayou, Nikhil Ghosh, and Bin Yu.
\newblock {LoRA}+: {Efficient} {Low} {Rank} {Adaptation} of {Large} {Models}, July 2024.
\newblock URL \url{http://arxiv.org/abs/2402.12354}.
\newblock arXiv:2402.12354 [cs].

\bibitem[Hinton et~al.(2015)Hinton, Vinyals, and Dean]{hinton_distilling_2015}
Geoffrey Hinton, Oriol Vinyals, and Jeff Dean.
\newblock Distilling the {Knowledge} in a {Neural} {Network}, March 2015.
\newblock URL \url{http://arxiv.org/abs/1503.02531}.
\newblock arXiv:1503.02531 [stat].

\bibitem[Hu et~al.(2021)Hu, Shen, Wallis, Allen-Zhu, Li, Wang, Wang, and Chen]{hu_lora_2021}
Edward~J. Hu, Yelong Shen, Phillip Wallis, Zeyuan Allen-Zhu, Yuanzhi Li, Shean Wang, Lu~Wang, and Weizhu Chen.
\newblock {LoRA}: {Low}-{Rank} {Adaptation} of {Large} {Language} {Models}, October 2021.
\newblock URL \url{http://arxiv.org/abs/2106.09685}.
\newblock arXiv:2106.09685 [cs].

\bibitem[Kaplun et~al.(2022)Kaplun, Malach, Nakkiran, and Shalev-Shwartz]{kaplun_knowledge_2022}
Gal Kaplun, Eran Malach, Preetum Nakkiran, and Shai Shalev-Shwartz.
\newblock Knowledge {Distillation}: {Bad} {Models} {Can} {Be} {Good} {Role} {Models}, March 2022.
\newblock URL \url{http://arxiv.org/abs/2203.14649}.
\newblock arXiv:2203.14649 [cs].

\bibitem[Kovaleva et~al.(2019)Kovaleva, Romanov, Rogers, and Rumshisky]{kovaleva_revealing_2019}
Olga Kovaleva, Alexey Romanov, Anna Rogers, and Anna Rumshisky.
\newblock Revealing the {Dark} {Secrets} of {BERT}, September 2019.
\newblock URL \url{http://arxiv.org/abs/1908.08593}.
\newblock arXiv:1908.08593 [cs].

\bibitem[Li et~al.(2025)Li, Meng, Zhang, Zhu, Wang, and Xu]{li_lorasuite_2025}
Yanan Li, Fanxu Meng, Muhan Zhang, Shiai Zhu, Shangguang Wang, and Mengwei Xu.
\newblock {LoRASuite}: {Efficient} {LoRA} {Adaptation} {Across} {Large} {Language} {Model} {Upgrades}, May 2025.
\newblock URL \url{http://arxiv.org/abs/2505.13515}.
\newblock arXiv:2505.13515 [cs].

\bibitem[Liang et~al.(2025)Liang, Tang, Zhou, Zhao, Shi, Zhao, Li, Wang, Schürholt, Borth, Bronstein, You, Wang, and Wang]{liang_drag-and-drop_2025}
Zhiyuan Liang, Dongwen Tang, Yuhao Zhou, Xuanlei Zhao, Mingjia Shi, Wangbo Zhao, Zekai Li, Peihao Wang, Konstantin Schürholt, Damian Borth, Michael~M. Bronstein, Yang You, Zhangyang Wang, and Kai Wang.
\newblock Drag-and-{Drop} {LLMs}: {Zero}-{Shot} {Prompt}-to-{Weights}, June 2025.
\newblock URL \url{http://arxiv.org/abs/2506.16406}.
\newblock arXiv:2506.16406 [cs].

\bibitem[Liu et~al.(2024)Liu, Wang, Yin, Molchanov, Wang, Cheng, and Chen]{liu_dora_2024}
Shih-Yang Liu, Chien-Yi Wang, Hongxu Yin, Pavlo Molchanov, Yu-Chiang~Frank Wang, Kwang-Ting Cheng, and Min-Hung Chen.
\newblock {DoRA}: {Weight}-{Decomposed} {Low}-{Rank} {Adaptation}, July 2024.
\newblock URL \url{http://arxiv.org/abs/2402.09353}.
\newblock arXiv:2402.09353 [cs].

\bibitem[Loshchilov and Hutter(2019)]{loshchilov_decoupled_2019}
Ilya Loshchilov and Frank Hutter.
\newblock Decoupled {Weight} {Decay} {Regularization}, January 2019.
\newblock URL \url{http://arxiv.org/abs/1711.05101}.
\newblock arXiv:1711.05101 [cs].

\bibitem[Mihaylov et~al.(2018)Mihaylov, Clark, Khot, and Sabharwal]{mihaylov_can_2018}
Todor Mihaylov, Peter Clark, Tushar Khot, and Ashish Sabharwal.
\newblock Can a {Suit} of {Armor} {Conduct} {Electricity}? {A} {New} {Dataset} for {Open} {Book} {Question} {Answering}, September 2018.
\newblock URL \url{http://arxiv.org/abs/1809.02789}.
\newblock arXiv:1809.02789 [cs].

\bibitem[MIRSKY(1960)]{mirsky_symmetric_1960}
L.~MIRSKY.
\newblock {SYMMETRIC} {GAUGE} {FUNCTIONS} {AND} {UNITARILY} {INVARIANT} {NORMS}.
\newblock \emph{The Quarterly Journal of Mathematics}, 11\penalty0 (1):\penalty0 50--59, January 1960.
\newblock ISSN 0033-5606.
\newblock \doi{10.1093/qmath/11.1.50}.
\newblock URL \url{https://doi.org/10.1093/qmath/11.1.50}.

\bibitem[OpenAI et~al.(2024)OpenAI, Achiam, Adler, Agarwal, Ahmad, Akkaya, Aleman, Almeida, Altenschmidt, Altman, Anadkat, Avila, Babuschkin, Balaji, Balcom, Baltescu, Bao, Bavarian, Belgum, Bello, Berdine, Bernadett-Shapiro, Berner, Bogdonoff, Boiko, Boyd, Brakman, Brockman, Brooks, Brundage, Button, Cai, Campbell, Cann, Carey, Carlson, Carmichael, Chan, Chang, Chantzis, Chen, Chen, Chen, Chen, Chen, Chess, Cho, Chu, Chung, Cummings, Currier, Dai, Decareaux, Degry, Deutsch, Deville, Dhar, Dohan, Dowling, Dunning, Ecoffet, Eleti, Eloundou, Farhi, Fedus, Felix, Fishman, Forte, Fulford, Gao, Georges, Gibson, Goel, Gogineni, Goh, Gontijo-Lopes, Gordon, Grafstein, Gray, Greene, Gross, Gu, Guo, Hallacy, Han, Harris, He, Heaton, Heidecke, Hesse, Hickey, Hickey, Hoeschele, Houghton, Hsu, Hu, Hu, Huizinga, Jain, Jain, Jang, Jiang, Jiang, Jin, Jin, Jomoto, Jonn, Jun, Kaftan, Kaiser, Kamali, Kanitscheider, Keskar, Khan, Kilpatrick, Kim, Kim, Kim, Kirchner, Kiros, Knight, Kokotajlo, Kondraciuk, Kondrich, Konstantinidis,
  Kosic, Krueger, Kuo, Lampe, Lan, Lee, Leike, Leung, Levy, Li, Lim, Lin, Lin, Litwin, Lopez, Lowe, Lue, Makanju, Malfacini, Manning, Markov, Markovski, Martin, Mayer, Mayne, McGrew, McKinney, McLeavey, McMillan, McNeil, Medina, Mehta, Menick, Metz, Mishchenko, Mishkin, Monaco, Morikawa, Mossing, Mu, Murati, Murk, Mély, Nair, Nakano, Nayak, Neelakantan, Ngo, Noh, Ouyang, O'Keefe, Pachocki, Paino, Palermo, Pantuliano, Parascandolo, Parish, Parparita, Passos, Pavlov, Peng, Perelman, Peres, Petrov, Pinto, Michael, Pokorny, Pokrass, Pong, Powell, Power, Power, Proehl, Puri, Radford, Rae, Ramesh, Raymond, Real, Rimbach, Ross, Rotsted, Roussez, Ryder, Saltarelli, Sanders, Santurkar, Sastry, Schmidt, Schnurr, Schulman, Selsam, Sheppard, Sherbakov, Shieh, Shoker, Shyam, Sidor, Sigler, Simens, Sitkin, Slama, Sohl, Sokolowsky, Song, Staudacher, Such, Summers, Sutskever, Tang, Tezak, Thompson, Tillet, Tootoonchian, Tseng, Tuggle, Turley, Tworek, Uribe, Vallone, Vijayvergiya, Voss, Wainwright, Wang, Wang, Wang, Ward,
  Wei, Weinmann, Welihinda, Welinder, Weng, Weng, Wiethoff, Willner, Winter, Wolrich, Wong, Workman, Wu, Wu, Wu, Xiao, Xu, Yoo, Yu, Yuan, Zaremba, Zellers, Zhang, Zhang, Zhao, Zheng, Zhuang, Zhuk, and Zoph]{openai_gpt-4_2024}
OpenAI, Josh Achiam, Steven Adler, Sandhini Agarwal, Lama Ahmad, Ilge Akkaya, Florencia~Leoni Aleman, Diogo Almeida, Janko Altenschmidt, Sam Altman, Shyamal Anadkat, Red Avila, Igor Babuschkin, Suchir Balaji, Valerie Balcom, Paul Baltescu, Haiming Bao, Mohammad Bavarian, Jeff Belgum, Irwan Bello, Jake Berdine, Gabriel Bernadett-Shapiro, Christopher Berner, Lenny Bogdonoff, Oleg Boiko, Madelaine Boyd, Anna-Luisa Brakman, Greg Brockman, Tim Brooks, Miles Brundage, Kevin Button, Trevor Cai, Rosie Campbell, Andrew Cann, Brittany Carey, Chelsea Carlson, Rory Carmichael, Brooke Chan, Che Chang, Fotis Chantzis, Derek Chen, Sully Chen, Ruby Chen, Jason Chen, Mark Chen, Ben Chess, Chester Cho, Casey Chu, Hyung~Won Chung, Dave Cummings, Jeremiah Currier, Yunxing Dai, Cory Decareaux, Thomas Degry, Noah Deutsch, Damien Deville, Arka Dhar, David Dohan, Steve Dowling, Sheila Dunning, Adrien Ecoffet, Atty Eleti, Tyna Eloundou, David Farhi, Liam Fedus, Niko Felix, Simón~Posada Fishman, Juston Forte, Isabella Fulford, Leo
  Gao, Elie Georges, Christian Gibson, Vik Goel, Tarun Gogineni, Gabriel Goh, Rapha Gontijo-Lopes, Jonathan Gordon, Morgan Grafstein, Scott Gray, Ryan Greene, Joshua Gross, Shixiang~Shane Gu, Yufei Guo, Chris Hallacy, Jesse Han, Jeff Harris, Yuchen He, Mike Heaton, Johannes Heidecke, Chris Hesse, Alan Hickey, Wade Hickey, Peter Hoeschele, Brandon Houghton, Kenny Hsu, Shengli Hu, Xin Hu, Joost Huizinga, Shantanu Jain, Shawn Jain, Joanne Jang, Angela Jiang, Roger Jiang, Haozhun Jin, Denny Jin, Shino Jomoto, Billie Jonn, Heewoo Jun, Tomer Kaftan, Łukasz Kaiser, Ali Kamali, Ingmar Kanitscheider, Nitish~Shirish Keskar, Tabarak Khan, Logan Kilpatrick, Jong~Wook Kim, Christina Kim, Yongjik Kim, Jan~Hendrik Kirchner, Jamie Kiros, Matt Knight, Daniel Kokotajlo, Łukasz Kondraciuk, Andrew Kondrich, Aris Konstantinidis, Kyle Kosic, Gretchen Krueger, Vishal Kuo, Michael Lampe, Ikai Lan, Teddy Lee, Jan Leike, Jade Leung, Daniel Levy, Chak~Ming Li, Rachel Lim, Molly Lin, Stephanie Lin, Mateusz Litwin, Theresa Lopez, Ryan
  Lowe, Patricia Lue, Anna Makanju, Kim Malfacini, Sam Manning, Todor Markov, Yaniv Markovski, Bianca Martin, Katie Mayer, Andrew Mayne, Bob McGrew, Scott~Mayer McKinney, Christine McLeavey, Paul McMillan, Jake McNeil, David Medina, Aalok Mehta, Jacob Menick, Luke Metz, Andrey Mishchenko, Pamela Mishkin, Vinnie Monaco, Evan Morikawa, Daniel Mossing, Tong Mu, Mira Murati, Oleg Murk, David Mély, Ashvin Nair, Reiichiro Nakano, Rajeev Nayak, Arvind Neelakantan, Richard Ngo, Hyeonwoo Noh, Long Ouyang, Cullen O'Keefe, Jakub Pachocki, Alex Paino, Joe Palermo, Ashley Pantuliano, Giambattista Parascandolo, Joel Parish, Emy Parparita, Alex Passos, Mikhail Pavlov, Andrew Peng, Adam Perelman, Filipe de Avila~Belbute Peres, Michael Petrov, Henrique Ponde de~Oliveira Pinto, Michael, Pokorny, Michelle Pokrass, Vitchyr~H. Pong, Tolly Powell, Alethea Power, Boris Power, Elizabeth Proehl, Raul Puri, Alec Radford, Jack Rae, Aditya Ramesh, Cameron Raymond, Francis Real, Kendra Rimbach, Carl Ross, Bob Rotsted, Henri Roussez,
  Nick Ryder, Mario Saltarelli, Ted Sanders, Shibani Santurkar, Girish Sastry, Heather Schmidt, David Schnurr, John Schulman, Daniel Selsam, Kyla Sheppard, Toki Sherbakov, Jessica Shieh, Sarah Shoker, Pranav Shyam, Szymon Sidor, Eric Sigler, Maddie Simens, Jordan Sitkin, Katarina Slama, Ian Sohl, Benjamin Sokolowsky, Yang Song, Natalie Staudacher, Felipe~Petroski Such, Natalie Summers, Ilya Sutskever, Jie Tang, Nikolas Tezak, Madeleine~B. Thompson, Phil Tillet, Amin Tootoonchian, Elizabeth Tseng, Preston Tuggle, Nick Turley, Jerry Tworek, Juan Felipe~Cerón Uribe, Andrea Vallone, Arun Vijayvergiya, Chelsea Voss, Carroll Wainwright, Justin~Jay Wang, Alvin Wang, Ben Wang, Jonathan Ward, Jason Wei, C.~J. Weinmann, Akila Welihinda, Peter Welinder, Jiayi Weng, Lilian Weng, Matt Wiethoff, Dave Willner, Clemens Winter, Samuel Wolrich, Hannah Wong, Lauren Workman, Sherwin Wu, Jeff Wu, Michael Wu, Kai Xiao, Tao Xu, Sarah Yoo, Kevin Yu, Qiming Yuan, Wojciech Zaremba, Rowan Zellers, Chong Zhang, Marvin Zhang, Shengjia
  Zhao, Tianhao Zheng, Juntang Zhuang, William Zhuk, and Barret Zoph.
\newblock {GPT}-4 {Technical} {Report}, March 2024.
\newblock URL \url{http://arxiv.org/abs/2303.08774}.
\newblock arXiv:2303.08774 [cs].

\bibitem[Phuong and Lampert(2021)]{phuong_towards_2021}
Mary Phuong and Christoph~H. Lampert.
\newblock Towards {Understanding} {Knowledge} {Distillation}, May 2021.
\newblock URL \url{http://arxiv.org/abs/2105.13093}.
\newblock arXiv:2105.13093 [cs].

\bibitem[Qwen et~al.(2025)Qwen, Yang, Yang, Zhang, Hui, Zheng, Yu, Li, Liu, Huang, Wei, Lin, Yang, Tu, Zhang, Yang, Yang, Zhou, Lin, Dang, Lu, Bao, Yang, Yu, Li, Xue, Zhang, Zhu, Men, Lin, Li, Tang, Xia, Ren, Ren, Fan, Su, Zhang, Wan, Liu, Cui, Zhang, and Qiu]{qwen_qwen25_2025}
Qwen, An~Yang, Baosong Yang, Beichen Zhang, Binyuan Hui, Bo~Zheng, Bowen Yu, Chengyuan Li, Dayiheng Liu, Fei Huang, Haoran Wei, Huan Lin, Jian Yang, Jianhong Tu, Jianwei Zhang, Jianxin Yang, Jiaxi Yang, Jingren Zhou, Junyang Lin, Kai Dang, Keming Lu, Keqin Bao, Kexin Yang, Le~Yu, Mei Li, Mingfeng Xue, Pei Zhang, Qin Zhu, Rui Men, Runji Lin, Tianhao Li, Tianyi Tang, Tingyu Xia, Xingzhang Ren, Xuancheng Ren, Yang Fan, Yang Su, Yichang Zhang, Yu~Wan, Yuqiong Liu, Zeyu Cui, Zhenru Zhang, and Zihan Qiu.
\newblock Qwen2.5 {Technical} {Report}, January 2025.
\newblock URL \url{http://arxiv.org/abs/2412.15115}.
\newblock arXiv:2412.15115 [cs].

\bibitem[Team et~al.(2024)Team, Riviere, Pathak, Sessa, Hardin, Bhupatiraju, Hussenot, Mesnard, Shahriari, Ramé, Ferret, Liu, Tafti, Friesen, Casbon, Ramos, Kumar, Lan, Jerome, Tsitsulin, Vieillard, Stanczyk, Girgin, Momchev, Hoffman, Thakoor, Grill, Neyshabur, Bachem, Walton, Severyn, Parrish, Ahmad, Hutchison, Abdagic, Carl, Shen, Brock, Coenen, Laforge, Paterson, Bastian, Piot, Wu, Royal, Chen, Kumar, Perry, Welty, Choquette-Choo, Sinopalnikov, Weinberger, Vijaykumar, Rogozińska, Herbison, Bandy, Wang, Noland, Moreira, Senter, Eltyshev, Visin, Rasskin, Wei, Cameron, Martins, Hashemi, Klimczak-Plucińska, Batra, Dhand, Nardini, Mein, Zhou, Svensson, Stanway, Chan, Zhou, Carrasqueira, Iljazi, Becker, Fernandez, Amersfoort, Gordon, Lipschultz, Newlan, Ji, Mohamed, Badola, Black, Millican, McDonell, Nguyen, Sodhia, Greene, Sjoesund, Usui, Sifre, Heuermann, Lago, McNealus, Soares, Kilpatrick, Dixon, Martins, Reid, Singh, Iverson, Görner, Velloso, Wirth, Davidow, Miller, Rahtz, Watson, Risdal, Kazemi,
  Moynihan, Zhang, Kahng, Park, Rahman, Khatwani, Dao, Bardoliwalla, Devanathan, Dumai, Chauhan, Wahltinez, Botarda, Barnes, Barham, Michel, Jin, Georgiev, Culliton, Kuppala, Comanescu, Merhej, Jana, Rokni, Agarwal, Mullins, Saadat, Carthy, Cogan, Perrin, Arnold, Krause, Dai, Garg, Sheth, Ronstrom, Chan, Jordan, Yu, Eccles, Hennigan, Kocisky, Doshi, Jain, Yadav, Meshram, Dharmadhikari, Barkley, Wei, Ye, Han, Kwon, Xu, Shen, Gong, Wei, Cotruta, Kirk, Rao, Giang, Peran, Warkentin, Collins, Barral, Ghahramani, Hadsell, Sculley, Banks, Dragan, Petrov, Vinyals, Dean, Hassabis, Kavukcuoglu, Farabet, Buchatskaya, Borgeaud, Fiedel, Joulin, Kenealy, Dadashi, and Andreev]{team_gemma_2024}
Gemma Team, Morgane Riviere, Shreya Pathak, Pier~Giuseppe Sessa, Cassidy Hardin, Surya Bhupatiraju, Léonard Hussenot, Thomas Mesnard, Bobak Shahriari, Alexandre Ramé, Johan Ferret, Peter Liu, Pouya Tafti, Abe Friesen, Michelle Casbon, Sabela Ramos, Ravin Kumar, Charline~Le Lan, Sammy Jerome, Anton Tsitsulin, Nino Vieillard, Piotr Stanczyk, Sertan Girgin, Nikola Momchev, Matt Hoffman, Shantanu Thakoor, Jean-Bastien Grill, Behnam Neyshabur, Olivier Bachem, Alanna Walton, Aliaksei Severyn, Alicia Parrish, Aliya Ahmad, Allen Hutchison, Alvin Abdagic, Amanda Carl, Amy Shen, Andy Brock, Andy Coenen, Anthony Laforge, Antonia Paterson, Ben Bastian, Bilal Piot, Bo~Wu, Brandon Royal, Charlie Chen, Chintu Kumar, Chris Perry, Chris Welty, Christopher~A. Choquette-Choo, Danila Sinopalnikov, David Weinberger, Dimple Vijaykumar, Dominika Rogozińska, Dustin Herbison, Elisa Bandy, Emma Wang, Eric Noland, Erica Moreira, Evan Senter, Evgenii Eltyshev, Francesco Visin, Gabriel Rasskin, Gary Wei, Glenn Cameron, Gus Martins,
  Hadi Hashemi, Hanna Klimczak-Plucińska, Harleen Batra, Harsh Dhand, Ivan Nardini, Jacinda Mein, Jack Zhou, James Svensson, Jeff Stanway, Jetha Chan, Jin~Peng Zhou, Joana Carrasqueira, Joana Iljazi, Jocelyn Becker, Joe Fernandez, Joost~van Amersfoort, Josh Gordon, Josh Lipschultz, Josh Newlan, Ju-yeong Ji, Kareem Mohamed, Kartikeya Badola, Kat Black, Katie Millican, Keelin McDonell, Kelvin Nguyen, Kiranbir Sodhia, Kish Greene, Lars~Lowe Sjoesund, Lauren Usui, Laurent Sifre, Lena Heuermann, Leticia Lago, Lilly McNealus, Livio~Baldini Soares, Logan Kilpatrick, Lucas Dixon, Luciano Martins, Machel Reid, Manvinder Singh, Mark Iverson, Martin Görner, Mat Velloso, Mateo Wirth, Matt Davidow, Matt Miller, Matthew Rahtz, Matthew Watson, Meg Risdal, Mehran Kazemi, Michael Moynihan, Ming Zhang, Minsuk Kahng, Minwoo Park, Mofi Rahman, Mohit Khatwani, Natalie Dao, Nenshad Bardoliwalla, Nesh Devanathan, Neta Dumai, Nilay Chauhan, Oscar Wahltinez, Pankil Botarda, Parker Barnes, Paul Barham, Paul Michel, Pengchong Jin,
  Petko Georgiev, Phil Culliton, Pradeep Kuppala, Ramona Comanescu, Ramona Merhej, Reena Jana, Reza~Ardeshir Rokni, Rishabh Agarwal, Ryan Mullins, Samaneh Saadat, Sara~Mc Carthy, Sarah Cogan, Sarah Perrin, Sébastien M.~R. Arnold, Sebastian Krause, Shengyang Dai, Shruti Garg, Shruti Sheth, Sue Ronstrom, Susan Chan, Timothy Jordan, Ting Yu, Tom Eccles, Tom Hennigan, Tomas Kocisky, Tulsee Doshi, Vihan Jain, Vikas Yadav, Vilobh Meshram, Vishal Dharmadhikari, Warren Barkley, Wei Wei, Wenming Ye, Woohyun Han, Woosuk Kwon, Xiang Xu, Zhe Shen, Zhitao Gong, Zichuan Wei, Victor Cotruta, Phoebe Kirk, Anand Rao, Minh Giang, Ludovic Peran, Tris Warkentin, Eli Collins, Joelle Barral, Zoubin Ghahramani, Raia Hadsell, D.~Sculley, Jeanine Banks, Anca Dragan, Slav Petrov, Oriol Vinyals, Jeff Dean, Demis Hassabis, Koray Kavukcuoglu, Clement Farabet, Elena Buchatskaya, Sebastian Borgeaud, Noah Fiedel, Armand Joulin, Kathleen Kenealy, Robert Dadashi, and Alek Andreev.
\newblock Gemma 2: {Improving} {Open} {Language} {Models} at a {Practical} {Size}, October 2024.
\newblock URL \url{http://arxiv.org/abs/2408.00118}.
\newblock arXiv:2408.00118 [cs].

\bibitem[Wang et~al.(2025)Wang, Ghosh, Cox, Antognini, Oliva, Feris, and Karlinsky]{wang_trans-lora_2025}
Runqian Wang, Soumya Ghosh, David Cox, Diego Antognini, Aude Oliva, Rogerio Feris, and Leonid Karlinsky.
\newblock Trans-lora: Towards data-free transferable parameter-efficient finetuning, May 2025.
\newblock URL \url{https://arxiv.org/abs/2405.17258}.
\newblock arXiv preprint arXiv:2405.17258.

\bibitem[Xu et~al.(2023)Xu, Xie, Qin, Tao, and Wang]{xu_parameter-efficient_2023}
Lingling Xu, Haoran Xie, Si-Zhao~Joe Qin, Xiaohui Tao, and Fu~Lee Wang.
\newblock Parameter-{Efficient} {Fine}-{Tuning} {Methods} for {Pretrained} {Language} {Models}: {A} {Critical} {Review} and {Assessment}, December 2023.
\newblock URL \url{http://arxiv.org/abs/2312.12148}.
\newblock arXiv:2312.12148 [cs].

\bibitem[Zellers et~al.(2019)Zellers, Holtzman, Bisk, Farhadi, and Choi]{zellers_hellaswag_2019}
Rowan Zellers, Ari Holtzman, Yonatan Bisk, Ali Farhadi, and Yejin Choi.
\newblock {HellaSwag}: {Can} a {Machine} {Really} {Finish} {Your} {Sentence}?, May 2019.
\newblock URL \url{http://arxiv.org/abs/1905.07830}.
\newblock arXiv:1905.07830 [cs].

\end{thebibliography}

\clearpage
\section{Appendix A. Additional Experimental Details}

\subsection{Datasets}
We evaluate Cross-LoRA on four widely used natural language understanding benchmarks, each targeting different aspects of reasoning and commonsense knowledge.  \\

\textbf{ARC-Challenge (ARC-C)}: A benchmark designed to test grade-school science reasoning ability. The dataset is composed of approximately 1.2k training examples and 1k evaluation questions, requiring multi-step reasoning beyond simple fact retrieval~\cite{clark_think_2018}. \\ 

\textbf{ARC-Easy (ARC-E)}: A companion benchmark to ARC-C with around 2.2k training examples. Unlike ARC-C, its questions are more straightforward, relying on direct factual recall. Together with ARC-C, it provides a balanced assessment of reasoning difficulty~\cite{clark_think_2018}.  \\

\textbf{OpenBookQA}: A benchmark focused on open-domain science knowledge with roughly 5k training questions. It requires combining information from a small science textbook (“open book”) with general commonsense reasoning~\cite{mihaylov_can_2018}. \\

\textbf{HellaSwag}: A large-scale commonsense inference benchmark with about 39k validation examples. It challenges models with adversarially filtered multiple-choice questions designed to be trivial for humans yet difficult for language models~\cite{zellers_hellaswag_2019}. \\ 

These datasets cover diverse reasoning challenges and scales, making them suitable to rigorously assess the generalization ability of Cross-LoRA.  
 
\subsection{LoRA Training Recipe}
We trained LoRA adapters~\cite{hu_lora_2021} across all four benchmarks, maintaining consistency in optimizer~\cite{loshchilov_decoupled_2019}, learning rate, and dropout, while adjusting training steps for dataset size.  
Table~\ref{tab:recipe} summarizes the settings used.

\begin{table}[H]
\centering
\small
\setlength{\tabcolsep}{4pt}
\resizebox{\columnwidth}{!}{%
\begin{tabular}{lcccc}
\toprule
\textbf{Setting} & \textbf{ARC-C} & \textbf{ARC-E} & \textbf{OBQA} & \textbf{HellaSwag} \\
\midrule
Batch Size     & 16 & 16 & 16 & 16 \\
Optimizer      & AdamW & AdamW & AdamW & AdamW \\
Learning Rate  & 1e-5 & 1e-5 & 1e-5 & 1e-5 \\
Training Steps & 600 & 600 & 600 & 300 \\
Weight Decay   & 0.1 & 0.1 & 0.1 & 0.1 \\
Max Grad Norm  & 1.0 & 1.0 & 1.0 & 1.0 \\
LoRA Alpha     & 32  & 32  & 32  & 32  \\
LoRA Dropout   & 0.1 & 0.1 & 0.1 & 0.1 \\
LoRA Bias      & None & None & None & None \\
LoRA Rank      & 16  & 16  & 16  & 16  \\
\bottomrule
\end{tabular}
}
\caption{LoRA training recipe across ARC-C, ARC-E, OBQA, and HellaSwag.}
\label{tab:recipe}
\end{table}

\subsection{Cross-LoRA Transferring}
For Cross-LoRA transfer, we applied adapters on a wide set of target modules: 

\begin{table}[ht]
\centering
\small
\setlength{\tabcolsep}{4pt}
\renewcommand{\arraystretch}{1.2}
\begin{tabularx}{\columnwidth}{Xcc}
\toprule
\textbf{Target Modules} & \textbf{Rank} & \textbf{LoRA Alpha} \\
\midrule
q\_proj, v\_proj, k\_proj, o\_proj, gate\_proj, up\_proj, down\_proj 
& 320 & 64 \\
\bottomrule
\end{tabularx}
\caption{Cross-LoRA transferring configuration.}
\label{tab:crosslora-config}
\end{table}

\section{Appendix B. Experiment Hardware}

We provide details of the hardware used for Cross-LoRA experiments. Batch experiments were carried out on NVIDIA Tesla V100 GPUs, while additional small-scale experiments were conducted on RTX 4090 for comparison.

Based on these results, we conservatively infer that Cross-LoRA should complete transfer in under 20 minutes on edge devices with 8GB GPU memory.

\begin{table}[ht]
\centering
\fontsize{8}{8}\selectfont
\setlength{\tabcolsep}{3pt}
\renewcommand{\arraystretch}{1.1}
\resizebox{\columnwidth}{!}{%
\begin{tabular}{lcc}
\toprule
\textbf{Setting} & \textbf{V100} & \textbf{RTX 4090} \\
\midrule
OS & Ubuntu 22.04.1 & Ubuntu 22.04.1 \\
CPU & \multicolumn{2}{c}{Intel(R) Xeon(R) Platinum 8352V CPU @ 2.10GHz} \\
GPU & NVIDIA Tesla V100 32GB & NVIDIA RTX 4090 24GB \\
CUDA Version & 11.8 & 11.8 \\
\shortstack{Cross-LoRA\\Transfer Time} & 349.287s & 564.043s \\
\shortstack{Cross-LoRA\\Memory Cost} & 5508MB & 2291MB \\
\bottomrule
\end{tabular}
}
\caption{Hardware platforms and Cross-LoRA transfer cost.}
\label{tab:hardware}
\end{table}

\section*{Appendix C. Detailed Derivation of Rank-Truncated SVD}
\label{appendix:truncation}

We analyze the approximation quality of rank-truncated singular value decomposition (SVD) as employed in Cross-LoRA.

\subsection*{Preliminaries}
Consider a weight matrix $W_0 \in \mathbb{R}^{3072 \times 3072}$ with singular value decomposition:
\[
W_0 = U \Sigma V^\top,
\]
where $\Sigma = \mathrm{diag}(\sigma_1, \sigma_2, \dots, \sigma_{3072})$ and $\sigma_1 \geq \sigma_2 \geq \cdots \geq 0$.  
The rank-$r$ truncated approximation of $W_0$ is:
\[
W_0^{(r)} = U_r \Sigma_r V_r^\top,
\]
where $U_r, V_r$ contain the top-$r$ singular vectors, and $\Sigma_r = \mathrm{diag}(\sigma_1, \dots, \sigma_r)$.  
By the Eckart--Young--Mirsky theorem~\cite{eckart_approximation_1936, mirsky_symmetric_1960}, $W_0^{(r)}$ minimizes the Frobenius norm error among all rank-$r$ matrices:
\[
\| W_0 - W_0^{(r)} \|_F^2 = \sum_{i=r+1}^{3072} \sigma_i^2.
\]

The fraction of Frobenius energy retained is defined as:
\[
\eta(r) = \frac{\sum_{i=1}^{r} \sigma_i^2}{\sum_{i=1}^{3072} \sigma_i^2}.
\]

\subsection*{LoRA Updates in the Truncated Subspace}
Given a LoRA update $\Delta W_s = B_s A_s$ with $\mathrm{rank}(\Delta W_s) \leq r$, it can be approximated in the truncated subspace as:
\[
\Delta W_s \approx U_r C V_r^\top, \quad C \in \mathbb{R}^{r \times r}.
\]

Projecting into the subspace yields:
\[
\widehat{\Delta W_s} = U_r (U_r^\top \Delta W_s V_r) V_r^\top,
\]
with corresponding Frobenius reconstruction error:
\[
E^2 = \|\Delta W_s - \widehat{\Delta W_s}\|_F^2 = \sum_{i=r+1}^{3072} \sigma_i^2.
\]

\subsection*{Numerical Illustration}
Empirical analyses suggest that low-rank approximations are sufficient for strong performance.  
Aghajanyan et al.~\cite{aghajanyan_intrinsic_2020} showed that tuning only 200 intrinsic dimensions retains roughly 90\% downstream accuracy, implying $\eta(200) \approx 0.9$.  
Moreover, spectral studies~\cite{kovaleva_revealing_2019} reveal rapid singular value decay in transformers, which can be approximated by a geometric decay $\sigma_i^2 \approx \sigma_1^2 \rho^{\,i}$, with $\rho \in [0.92, 0.97]$.

Under this assumption:
\[
\eta(r) \approx 1 - \rho^r.
\]
For $r = 320$ and $\rho = 0.94$:
\[
\eta(320) \approx 1 - (0.94)^{320} \approx 1 - e^{-19.8} \approx 0.999999998.
\]
Even for $\rho = 0.97$:
\[
\eta(320) \approx 1 - (0.97)^{320} \approx 0.99994.
\]

Thus, in typical transformer layers, rank-$320$ truncation retains at least 99\% of the Frobenius norm energy. The reconstruction error is bounded by:
\[
E^2 = (1 - \eta(320)) \cdot \|W_0\|_F^2, \quad \text{so } E \leq \sqrt{0.001} \approx 0.031.
\]

\subsection*{Conclusion}
These results justify that rank-$320$ truncation preserves nearly all semantically meaningful structure in $W_0$.  
Given that LoRA updates themselves are rank-$16$ or lower, the approximation error is negligible in practice.  
This explains why Cross-LoRA achieves performance comparable to directly trained LoRA adapters.

\end{document}